\documentclass[sigconf]{acmart}

\copyrightyear{2025} 
\acmYear{2025} 
\setcopyright{cc}
\setcctype{by}
\acmConference[KDD '25]{Proceedings of the 31st ACM SIGKDD Conference on Knowledge Discovery and Data Mining V.2}{August 3--7, 2025}{Toronto, ON, Canada}
\acmBooktitle{Proceedings of the 31st ACM SIGKDD Conference on Knowledge Discovery and Data Mining V.2 (KDD '25), August 3--7, 2025, Toronto, ON, Canada}
\acmDOI{10.1145/3711896.3736877}
\acmISBN{979-8-4007-1454-2/2025/08}

\AtBeginDocument{
  }

\usepackage[linesnumbered,ruled,vlined]{algorithm2e}
\usepackage[utf8]{inputenc} 
\usepackage[T1]{fontenc} 
\usepackage{hyperref}   
\usepackage{url}      
\usepackage{booktabs} 
\usepackage{tabularx}
\usepackage{amsfonts}  
\usepackage{nicefrac} 
\usepackage{microtype} 
\usepackage{xcolor}    
\usepackage{graphicx}
\usepackage{multicol}
\usepackage{array}
\usepackage{svg}
\usepackage{multirow}
\usepackage{enumitem}
\usepackage{ dsfont }
\usepackage{amsthm}
\usepackage{adjustbox}
\theoremstyle{boldthm}
\newtheorem{theorem}{Theorem}

\newcommand{\vect}[1]{\boldsymbol{#1}}

\def\header{\vspace{1mm} \noindent}

\begin{document}

\title{Chi-Square Wavelet Graph Neural Networks for Heterogeneous Graph Anomaly Detection}

\author{Xiping Li}
\authornote{This work was done during an internship at CUHK and Tencent.}
\affiliation{
  \institution{The Chinese University of Hong Kong}
  \city{Hong Kong SAR}
    \country{China}
}
\email{lihsiping@gmail.com}

\author{Xiangyu Dong}
\affiliation{
  \institution{The Chinese University of Hong Kong}
  \city{Hong Kong SAR}
    \country{China}
}
\email{xydong@se.cuhk.edu.hk}

\author{Xingyi Zhang}
\affiliation{
  \institution{Mohamed bin Zayed University of
 Artificial Intelligence}
  \city{Abu Dhabi}
    \country{United Arab Emirates
}
}
\email{xingyi.zhang@mbzuai.ac.ae}

\author{Kun Xie}
\affiliation{
  \institution{The Chinese University of Hong Kong}
  \city{Hong Kong SAR}
    \country{China}
}
\email{kxie@se.cuhk.edu.hk}

\author{Yuanhao Feng}
\affiliation{
  \institution{Tencent, WeChat Pay}
  \city{Shenzhen}
    \country{China}
}
\email{danielfeng@tencent.com}

\author{Bo Wang}
\affiliation{
  \institution{Tencent, WeChat Pay}
  \city{Shenzhen}
    \country{China}
}
\email{pollowang@tencent.com}

\author{Guilin Li}
\affiliation{
  \institution{Tencent, WeChat Pay}
  \city{Shenzhen}
    \country{China}
}
\email{guilinli@tencent.com}

\author{Wuxiong Zeng}
\affiliation{
  \institution{Tencent, WeChat Pay}
  \city{Shenzhen}
    \country{China}
}
\email{philzeng@tencent.com}

\author{Xiujun Shu}
\affiliation{
  \institution{Tencent, WeChat Pay}
  \city{Shenzhen}
    \country{China}
}
\email{xiujunshu@tencent.com}

\author{Sibo Wang}
\authornote{Corresponding author}
\affiliation{
  \institution{The Chinese University of Hong Kong}
  \city{Hong Kong SAR}
    \country{China}
}
\email{swang@se.cuhk.edu.hk}

\renewcommand{\shortauthors}{Xiping Li et al.}

\begin{abstract}
\label{sec:abstract}

Graph Anomaly Detection (GAD) in heterogeneous networks presents unique challenges due to node and edge heterogeneity. Existing Graph Neural Network (GNN) methods primarily focus on homogeneous GAD and thus fail to address three key issues: (C1) Capturing abnormal signal and rich semantics across diverse meta-paths; (C2) Retaining high-frequency content in HIN dimension alignment; and (C3) Learning effectively from difficult anomaly samples with class imbalance. To overcome these, we propose ChiGAD, a spectral GNN framework based on a novel Chi-Square filter, inspired by the wavelet effectiveness in diverse domains. Specifically, ChiGAD consists of: (1) Multi-Graph Chi-Square Filter, which captures anomalous information via applying dedicated Chi-Square filters to each meta-path graph; (2) Interactive Meta-Graph Convolution, which aligns features while preserving high-frequency information and incorporates heterogeneous messages by a unified Chi-Square Filter; and (3) Contribution-Informed Cross-Entropy Loss, which prioritizes difficult anomalies to address class imbalance. Extensive experiments on public and industrial datasets show that ChiGAD outperforms state-of-the-art models on multiple metrics. Additionally, its homogeneous variant, ChiGNN, excels on seven GAD datasets, validating the effectiveness of Chi-Square filters. Our code is available at https://github.com/HsipingLi/ChiGAD.

\end{abstract}



\maketitle

\section{Introduction}
\label{sec:introduction}

Graph Anomaly Detection (GAD) \cite{XGBGraph} has emerged as a critical research focus of graph learning
, with vital applications ranging from identifying risky merchants \cite{risky_merchants_1} to detecting fake news, \cite{fake_news_1, kpg}, and malicious users \cite{malicious_users_1, malicious_users_2}. Graph Neural Networks (GNNs) \cite{GCN, GAT} have become the preferred solution for GAD tasks \cite{BWGNN, GHRN}, given their inherent ability to process graph-structured data. Since node-level GAD is fundamentally a binary node classification task, thorough assessments of various traditional GNN models \cite{GCN, GraphSAGE} have been conducted. Recent spectral GNNs \cite{BWGNN, GHRN, AMNet}, especially those employing wavelet filters \cite{BWGNN, GAD_wavelet_1, GAD_wavelet_2}, have shown great superiority in homogeneous GAD tasks. Yet, these approaches invariably assume unified node and edge types of fraud graphs, ignoring the scenarios characterized by Heterogeneous Information Networks (HINs). Despite the successful application of GNN-based models in various HINs, including social network dynamics \cite{HIN_social_1, HIN_social_2}, link prediction \cite{HIN_linkpred_1, HIN_linkpred_2}, and recommendations \cite{HIN_recsys_1, HIN_recsys_2, HIN_recsys_3, HIN_recsys_4, HIN_recsys_5}, anomaly detection in heterogeneous networks remains understudied. Considering the significance of GAD in critical HINs such as financial networks, a specialized framework for heterogeneous GAD is urgently needed.

Learning on HINs is inherently challenging due to both the node and edge heterogeneity \cite{HetGNN}. Recently, a series of meta-path-based GNNs \cite{HAN, MAGNN, RSHN, HetGNN, HGT} and multi-graph filters \cite{PSHGCN, MGNN} have been proposed to address the issues arising from heterogeneity. However, it is worth noting that they overlook the following critical challenges due to not being specifically designed for GAD tasks. Particularly, 
\textbf{(C1) Capturing abnormal signals and rich semantics across diverse meta-paths:} In GAD applications such as fraud detection, meta-paths exhibit distinct frequency characteristics crucial for anomaly identification. For instance, in a financial network, transaction-based meta-paths may show irregular bursts, while social interactions remain steady. A generic filter cannot capture such differences across frequency bands. Previous works \cite{MGNN, PSHGCN} adopt a unified signal processing framework, thereby overlooking the nuanced frequency energy distributions inherent in different meta-path graphs. 
\textbf{(C2) Retaining high-frequency content in HIN dimension alignment:} 
The retention of high-frequency content is essential to avert over-smoothing and performance degradation in homogeneous GAD tasks \cite{BWGNN, GHRN}; additionally, the node heterogeneity inherent in HIN engenders dimensional disparity, compelling nodes to depend on a prior dimensional alignment technique for feasible representation learning within heterogeneous graphs \cite{HetGNN}. Nevertheless, the risky potential for the alignment method to induce loss of high-frequency content within these signals remains an unexplored domain since they directly process graph signals.
\textbf{(C3) Learning effectively from difficult anomaly samples with class imbalance:} In GAD tasks, the inherent class imbalance poses significant challenges. This imbalance makes it difficult for models to learn subtle anomaly characteristics. While existing studies \cite{RQGNN} have acknowledged this challenge, a comprehensive strategy addressing class imbalance while effectively learning from rare and difficult anomaly samples is still lacking.

Inspired by the demonstrated effectiveness of wavelet in various domains \cite{appli_wavelet_1, appli_wavelet_2}, especially in homogeneous GAD \cite{BWGNN, GAD_wavelet_1, GAD_wavelet_2}, we first employ the Chi-Square distribution to design a novel graph filter, namely, \textbf{Chi-Square filter}. This novel graph wavelet filter not only inherits the proven wavelet properties crucial for GAD tasks \cite{BWGNN, GAD_wavelet_1, GAD_wavelet_2}, but also possesses a unique property of additivity compared to other wavelets, which makes Chi-Square naturally suitable for collaborative graph representation learning across diverse meta-paths within HIN, as elaborated in Section \ref{sec: wavelet properties}. 

Furthermore, motivated by the aforementioned three crucial challenges, we introduce a novel framework for heterogeneous graph anomaly detection, namely \textbf{ChiGAD}, which takes our proposed Chi-Square filter as a basic component and is comprised of three essential components: 
\textbf{(1) Multi-Graph Chi-Square Filter}: It processes different node types independently. For each type of nodes, it leverages the additivity of Chi-Square to collaboratively capture fine-grained semantics across diverse meta-paths by assigning the most aligned Chi-Square filters according to the spectral energy distributions of representative meta-path graphs; 
\textbf{(2) Interactive Meta-Graph Convolution}: It addresses representation dimension disparities in heterogeneous graphs by using a linear layer to align feature dimensions, justified by its ability to preserve high-frequency components to resist over-smoothing, which is helpful to heterogeneous GAD tasks, as proven in Section \ref{sec: method II interactive xxx} in theory and analyzed in Appendix \ref{app: experiments for comparison of alignment method}. It aligns node representations across types and employs a unified Chi-Square filter on a transformed homogeneous graph to learn final nodal representations, integrating diverse frequency bands; \textbf{(3) Contribution-Informed Cross-Entropy Loss}: It assigns higher weights to anomalies based on their contributions, ensuring the model prioritizes difficult samples while maintaining focus on the scarcity of anomalous data. 

In summary, our key contributions are as follows:
\begin{itemize}[topsep=0.5mm, partopsep=0pt, itemsep=0pt, leftmargin=10pt] 
\item To the best of our knowledge, we are the first to use Chi-Square wavelets for spectral GNNs. In addition to properties possessed by general wavelets, the unique additivity of Chi-Square wavelet enables collaborative learning across multiple meta-paths, making them seamlessly suited for heterogeneous GAD.
\item We propose ChiGAD, a novel spectral GNN framework for heterogeneous GAD, which captures rich semantics across diverse meta-paths and broad frequency bands, retains the high-frequency content in alignment, and effectively learns from difficult anomalies while mitigating the class-imbalance problem.
\item We provide theoretical analysis and empirical validation of high-frequency retention during dimensional alignment, which is crucial for mitigating over-smoothing in GAD tasks. Both theoretical findings in Theorem \ref{thm: 1} and experimental results shown in Appendix \ref{app: experiments for comparison of alignment method} indicate that linear layer is a good choice for dimensional alignment in heterogeneous GAD.
\item Experimental results on both industrial and public heterogeneous datasets demonstrate the effectiveness of ChiGAD in heterogeneous GAD tasks, showing significant improvements over state-of-the-art baselines across diverse metrics, including AUROC, AUPRC, F1-macro, and Recall.
\item We develop ChiGNN, a homogeneous variant of ChiGAD, which achieves superior performance on seven public datasets, further validating the effectiveness of the Chi-Square filter. 
 \end{itemize}

\section{Related Work}
\label{sec: related work}

\subsection{GNN-based Graph Anomaly Detection}
GNN-based models are popular for node-level GAD in homogeneous graphs \cite{XGBGraph, smoothgnn, spacegnn}. However, their effectiveness is limited by the low-pass filtering of Graph Convolutional Networks \cite{GCN, GHRN, BWGNN} and the heterophily phenomenon \cite{GHRN, GAGA}, where anomalies connect to normal nodes. 
Several GNN-based anomaly detectors \cite{GAGA, AMNet, BWGNN, GHRN} have been proposed to address these challenges. Specifically, BWGNN \cite{BWGNN} incorporates Beta wavelet band-pass filters to detect graph anomalies, leveraging the right-shift phenomenon in spectral energy. XGBGraph \cite{XGBGraph} aggregates neighborhood features within an XGBoost model to capture both structural and feature cues. GHRN \cite{GHRN} mitigates heterophily by pruning edges identified via a high-frequency indicator, while GAGA \cite{GAGA} improves fraud detection in low-homophily graphs through group aggregation and learnable encoding. ConsisGAD \cite{ConsisGAD} addresses limited supervision with pseudo-label generation. Yet, these models invariably overlook heterogeneous scenarios that better reflect real-world conditions.

\subsection{Heterogeneous Graph Learning.}
Diverse investigations \cite{HAN, MAGNN, RSHN, HetGNN, MGNN, PSHGCN} have been introduced to enable graph representation learning in heterogeneous networks, addressing the limitations of traditional GNN-based models. Meta-path-based models provide foundational solutions. Specifically, HAN \cite{HAN} uses a hierarchical attention mechanism combined with multiple meta-paths to capture node-level and semantic-level importance in HINs. MAGNN \cite{MAGNN} enhances this approach by encoding all meta-path information rather than just endpoints, addressing the limitation of ignoring intermediate nodes. HetGNN \cite{HetGNN} leverages random walks with restart and Bi-LSTM \cite{Bi-LSTM} to aggregate node features across node types. Recently, spectral GNN extensions to HINs have emerged. MGNN \cite{MGNN} proposes a polynomial multi-graph filter where each monomial term serves as a non-commutative multiplication of graph shifting operators. PSHGCN \cite{PSHGCN} introduces a positive semi-definite constraint to enhance model performance. Yet, these approaches are not specifically designed for GAD tasks. 

Multi-view clustering also models relations across several representations of the same samples. PRTL combines low-rank tensor regularization with a hyper-Laplacian graph to capture higher-order sample relations~\cite{PRTL}, while HSIMTC uses a hierarchical tensor to learn clustering affinities when some views are missing~\cite{HSIMTC}. ChiGAD uses meta-path graphs to represent distinct relations for supervised node-level anomaly detection.

\section{Preliminaries}
\label{sec:preliminary}

\subsection{Heterogeneous Information Networks}\label{subsec:hin}
A heterogeneous information network (HIN) is defined as a graph $G_M=(\mathcal{V},\mathcal{A},\mathcal{X},O_V, R_E)$, where $\mathcal{V}=\bigcup_{o=1}^{|O_V|}\{V_o\}$ represents multiple types of nodes, each enriched with attributes  $\mathcal{X}=\{\vect{X}_o\}_{o=1}^{|O_V|}$; $\mathcal{A}=\{\vect{A}_r\}_{r=1}^{|R_E|}$ includes the adjacency matrices corresponding to diverse relations, where $(\vect{A}_r)_{ij}=1$ if there exists an edge between nodes $i,j\in \mathcal{V}$ under relation $r$, otherwise $(\vect{A}_r)_{ij}=0$; 
$O_V$ and $R_E$ respectively denote the sets of node types and relation types, and $|O_V|+|R_E|>2$. 
For HINs, it is usually observed that there is a dimensional disparity in node attributes $\vect{X}_o\in \mathbb{R}^{|V_o|\times d_o}$, i.e. $d_i \neq d_j$ for $i\neq j,~i,j \in O_V$; in addition, $\vect{A}_r \in \mathbb{R}^{|\mathcal{V}|\times |\mathcal{V}|}$ for $r=1,2,\cdots,|R_E|$.

\subsection{Meta-Path Graphs}\label{subsec:mpg}

The meta-path is a path pattern defined on a given heterogeneous information network $G_M$, describing an alternate sequence of node types and edge types. A meta-path $P$ of length $l$ can be defined as:
\begin{equation}\label{eq: P=o-r-o-r-o}
    P = o_1 \xrightarrow{r_1} o_2 \xrightarrow{r_2} \cdots \xrightarrow{r_{l}} o_{l+1}, 
\end{equation}
where $\{o_i\}_{i=1}^{l+1}$ and $\{r_i\}_{i=1}^{l}$ are collections of node types and relations, respectively. In addition, a meta-path graph $G_P$ is a subgraph extracted from the meta-graph $G_M$ based on $P$ defined in Equation \ref{eq: P=o-r-o-r-o}, containing nodes and edges that conform to $P$.

The meta-path reveals the high-level semantic information hidden in HINs. For instance, in an academic network, the meta-path:
\begin{equation}\label{eq:a-p-a}
    author \xrightarrow{compose} paper \xrightarrow{composed~by} author  \nonumber
\end{equation}
captures co-authoring semantics among authors. The diversity of meta-path composition is an inherent property of HINs, and effectively mining rich information across meta-paths way poses challenges for HIN-related tasks \cite{HAN}.

\subsection{Dynamics of Information Diffusion}\label{subsec:diffusion}

Meta-paths capture complex relations between nodes in HINs. Given an HIN $G_M$, each meta-path $P$ as defined in Equation \ref{eq: P=o-r-o-r-o} varies based on both the composition and length of its sequence $r_1,r_2,\cdots,r_l$. Following prior work \cite{MGNN}, information diffusion along a meta-path $P$ refers to an alternate execution of message propagation and aggregation, formally expressed as:
\begin{equation}\label{eq:x=sssx}
    \vect{X}^{'} = \vect{S}_{r_l}\cdots \vect{S}_{r_2}\vect{S}_{r_1}\vect{X}, \nonumber
\end{equation}
where $\vect{S}_{r_i}$ is the graph shift operator of the edge-type subgraph $G_{r_i}=(V,\vect{A}_{r_i},\vect{X})$, for $i=1,\cdots,l$. Notably, given $\vect{S}_{{r}_i}=\vect{A}_{{r}_i}$, the multiplication $\vect{A}_{{r}_{l}} \cdots \vect{A}_{{r}_2}\vect{A}_{{r}_1}$ is the diffussion process of a meta-path graph $G_{P}$ based on $P$. The flexible choice of graph shift operator in the design of spectral GNNs allows us to view message propagation and aggregation within spatial domains as a special case of the spectral perspective, bridging these two domains and providing an intuitive grasp of multi-graph filters.

\subsection{Multi-Graph Filters}\label{subsec:multigrpahfilters}

A multi-graph spectral GNN \cite{MGNN, PSHGCN} is formulated as a multivariate polynomial $H(\vect{S}_1, \vect{S}_2, \cdots, \vect{S}_m)$, where $\vect{S}_1, \cdots, \vect{S}_m$ are graph shift operators, either adjacency matrices or Laplacian matrices of $m$ homogeneous graphs $G_1, ..., G_m$. Formally, $H$ is defined as:
\begin{equation}\label{eq: H}
    H(\vect{S}_1, \cdots, \vect{S}_m) = \sum_{i=1}^{m}f_i(\vect{S}_i) + f_{mv}(\vect{S}_1,\cdots,\vect{S}_m),
\end{equation}
where the first part processes signals within each graph $G_i$ independently using univariate polynomial $f_i,~i=1,\cdots,m$, while the second part $f_{mv}$ incorporates multivariate monomials, e.g. $\vect{S}_3\vect{S}_2\vect{S}_1$. This multi-graph filter degenerates to an ordinary spectral GNN for homogeneous graphs when $m=1$.

\subsection{High-frequency Area}\label{subsec: S_high intro}
Spectral GNNs leverage the Laplacian spectrum of input graphs as graph shift operators to collect spectral information: For any given graph $G=(V,\vect{A},\vect{X})$, the Laplacian spectrum can be represented by its Laplacian matrix $\vect{L}=\vect{D}-\vect{A}$, where $\vect{D}$ is the diagonal degree matrix and $\vect{A}$ is the adjacency matrix. The spectral decomposition is defined as $\vect{L} = \vect{U}\vect{\Lambda} \vect{U}^{\top}$, where $\vect{U}$ contains normalized eigenvectors as columns and $\vect{\Lambda}$ is a diagonal matrix of eigenvalues. Following previous work \cite{BWGNN}, the high-frequency area can be computed by
\begin{equation}\label{eq: S_high}
    S_{high} = \frac{\vect{x}^{\top}\vect{L}\vect{x}}{\vect{x}^{\top}\vect{x}} \in \mathcal{R}^1,
\end{equation}
where $\vect{x}$ serves as a graph signal for nodes in $G$, and $\vect{L}$ is the graph Laplacian matrix. 
According to existing works \cite{RQGNN, GHRN, BWGNN}, $S_{high}$ reflects both the high-frequency content and the smoothness of graph signal $\vect{x}$ from spectral and spatial domains, respectively, providing valuable properties to tackle challenges mentioned in Section~\ref{sec:introduction}. Moreover, its computation efficiency, as it avoids eigenvalue calculations \cite{BWGNN}, makes it an ideal metric for profiling the spectral energy distribution of signal $\vect{x}$ on graph $G$.

\subsection{Heterogenous Graph Anomaly Detection}\label{sec:HGAD}
Given a heterogeneous graph $G_M=(\mathcal{V},\mathcal{A},\mathcal{X},O_V, R_E)$ and a target node type $o_t \in O_V$, let $V_{o_t}^{an}, V_{o_t}^{no} \subsetneq V_{o_t}$ be two disjoint node subsets ($V_{o_t}^{an}\bigcap V_{o_t}^{no}=\varnothing$) representing anomalous and normal nodes, respectively. The GAD task aims to identify nodes belonging to $V_{o_t}^{an}$, which typically exhibit atypical patterns regarding structures or attributes \cite{XGBGraph}. In addition, unlike general binary classification, GAD faces a severe class imbalance challenge, where $|V_{o_t}^{an}| \ll |V_{o_t}^{no}|$ \cite{XGBGraph, RQGNN}. Building on the foregoing discussion, this study aims to devise a novel spectral GNN to address the key challenges outlined in Section \ref{sec:introduction}, specifically tailored for node-level GAD in heterogeneous networks. We present the detailed framework in Section \ref{sec:method}.

\section{ChiGAD}
\label{sec:method}
\begin{figure*}[t]
\centering
\begin{minipage}{\textwidth}
\hspace{4mm}
 \includegraphics[width=0.94\textwidth]{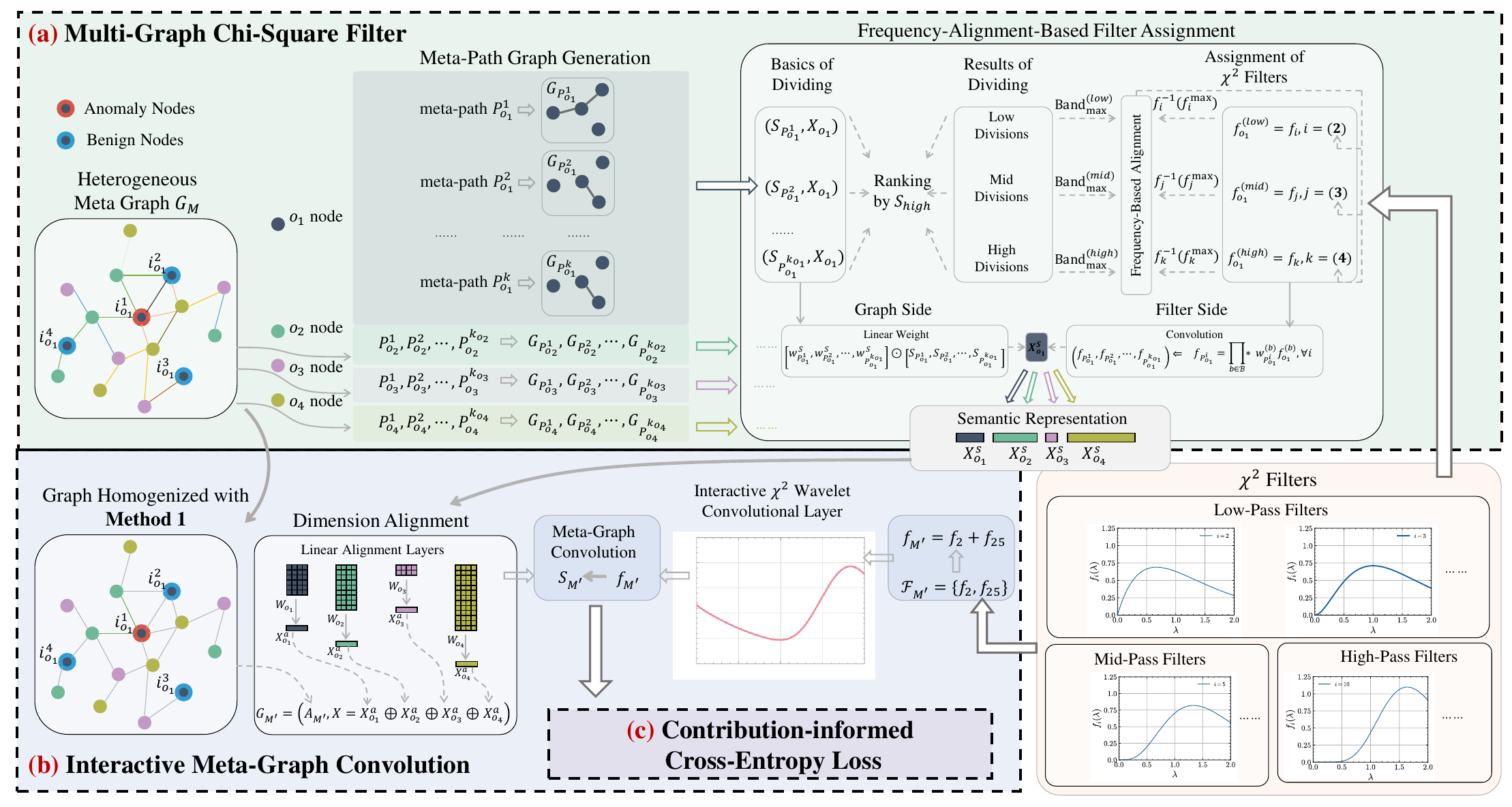}
\vspace{-5mm}
\caption{Illustration of the ChiGAD framework, comprising three modules: (a) Multi-Graph Chi-Square Filter; (b) Interactive Meta-Graph Convolution; (c) Contribution-Informed Cross-Entropy Loss.}
\label{fig: framework}
\end{minipage} \vspace{-4mm}
\end{figure*}

Wavelet filters have shown superiority in various tasks \cite{appli_wavelet_1, appli_wavelet_2} including GAD \cite{BWGNN, GAD_wavelet_1, GAD_wavelet_2}, which motivates our proposal of the Chi-Square filter. In this section, we first introduce the Chi-Square filter, which serves as the fundamental component of our ChiGAD framework, proving its wavelet properties and highlighting its unique characteristics.
Based on these preliminaries, we present ChiGAD as a comprehensive solution to the three challenges outlined in Section \ref{sec:introduction}. ChiGAD comprises three key components: (1) Multi-Graph Chi-Square Filter, (2) Interactive Meta-Graph Convolution, and (3) Contribution-Informed Cross-Entropy Loss. The overall architecture of ChiGAD is depicted in Figure \ref{fig: framework}.

\subsection{Wavelet Properties of Chi-Square}
\label{sec: wavelet properties}
Recent research has demonstrated the remarkable efficacy of graph wavelet \cite{Hammond} among various graph filters \cite{AMNet, BernNet, MGNN, BWGNN, PSHGCN}, particularly in domains such as anomaly detection, signal processing and computer vision \cite{BWGNN, appli_wavelet_1,appli_wavelet_2,GAD_wavelet_1,GAD_wavelet_2}.
Inspired by these studies, we introduce a novel graph wavelet based on the Chi-Square distribution. In this section, we demonstrate that the Chi-Square distribution not only satisfies the theoretical requirements of a graph wavelet but also offers a unique additivity property. This additivity, complementing the established advantages of conventional wavelets, proves particularly valuable for heterogeneous GAD.

Consider the Chi-Square distribution with \( n \) degrees of freedom. Its probability density function is defined as
$$
    f(x) = \frac{1}{2^{\frac{n}{2}} \Gamma\left(\frac{n}{2}\right)} x^{\frac{n}{2} - 1} e^{-\frac{x}{2}}, \quad x > 0,
$$
where \( e \) is the natural logarithm base and \( \Gamma(\cdot) \) is the Gamma function \cite{Gamma}.
While $f(x)$ is not a linear polynomial, it can be well approximated by 3-order Chebyshev polynomials \cite{Cheby_1, Cheby_2}, as validated by empirical results. In this work, we introduce a scaling factor $m$, and the Chi-Square filter with parameter $i$ is defined as:
\begin{align}\label{eq: f_i}
    f_i(w) &= \frac{1}{S_i}\frac{1}{2^{\frac{n}{2}} \Gamma(\frac{n}{2})} \left(\frac{w}{m}\right)^{\frac{n}{2}-1} e^{-\frac{w}{2m}} \nonumber \\
    &= \frac{1}{S_i}\frac{1}{2^{i} \Gamma(i)} \left(w (i + 1)\right)^{i - 1} e^{-\frac{w (i + 1)}{2}}, \quad i=1,2,\cdots, 
\end{align}
where $n=2i$,$m=\frac{1}{i+1}$, and $S_i=\int_0^{2} \frac{1}{2^{i} \Gamma(i)} \left(w (i + 1)\right)^{i - 1} e^{-\frac{w (i + 1)}{2}}dw$ regularizes $f_i$ into a probability density function in $[0,2]$, which is the full frequency range \cite{BWGNN}. Such scaling enhances the expressiveness of $f_i$ in processing signals of different frequencies while retaining the crucial properties of $f_i$, which are detailed below:

\noindent
\textbf{(1) Admissibility Condition:}
\begin{equation}\label{eq:admissibility}
    \int_0^{\infty} \frac{|f_i(w)|^2}{w}dw<\infty. \nonumber
\end{equation}
It ensures filter $f_i$ satisfies wavelet requirements and maintains band-pass property \cite{Hammond}. Proof is shown in Appendix \ref{sec: proof of admissibility condition}. 

\noindent
\textbf{(2) Broad Band Coverage:} Ideal filters $\{f_i\}_{i=1}^{\infty}$ are expected to cover diverse frequency bands. To this end, empirical results shown in Appendix \ref{sec: broad coverage} verify the broad coverage of both $\{\mathds{E}(\textbf{X}_i)~|~\mathds{E}(\textbf{X}_i)=\frac{1}{S_i}\frac{m}{n}=\frac{1}{S_i}\frac{n}{m}=\frac{1}{S_i}\frac{2i}{i+1}\}_{i\geq 1}$ and $\{f_{i}^{-1}(f_i^{\max}) ~|~ f_{i}^{-1}(f_i^{\max})=\arg\max\limits_{x}$\\$ f_i(x)\}_{i\geq 1}$, which are the expectation of $\textbf{X}_i \sim f_i(x)$ and the most highlighted frequency for any $i\geq 1$, respectively.

\noindent
\textbf{(3) Spectral Locality:} For  $\textbf{X}_i \sim f_i(x),~\forall i=2,3,\cdots$,
\begin{align}\label{spectrallocal}
    \mathds{E}(\textbf{X}_i)=\frac{1}{S_i} \frac{2i}{i+1}, \nonumber
    ~~\mathds{V}(\textbf{X}_i)=\frac{1}{S_i} \frac{2i}{(i+1)^2}. \nonumber
\end{align}
As $f_i$ is regularized to a distribution in $[0,2]$, we have $\mathds{E}(\textbf{X}_i) \in [0,2]$ and $\mathds{V}(\textbf{X}_i)\rightarrow 0$ when $i \rightarrow \infty$ .

\noindent
\textbf{(4) Spatial Locality:} When using a $d$-order Chebyshev polynomial,
$f_i(L)$ is an $(i-1+d)$-order polynomial of the Laplacian matrix $L$. Following Lemma 5.2 in existing research \cite{Hammond}, signal effects are localized within $(i-1+d)$-hop neighbors (where $d=3$).

\subsection{Multi-Graph Chi-Square Filter}
\label{sec: method: parallel multi-graph filter}

\header{\bf Meta-Path Graph Generation.}  
Node heterogeneity, characterized by diverse node types, presents several challenges for graph representation learning, such as attribute dimension disparity \cite{HetGNN}. Our approach addresses this by learning representations for each node type independently using Chi-Square filters, a strategy that not only mitigates dimensional disparity but also enables effective representation capture, leveraging the powerful properties of the Chi-Square filter, as shown in our experiments.

To capture rich semantic relationships, we first generate a set of meta-paths \( P \), as defined in Equation \ref{eq: P=o-r-o-r-o}, ensuring that:
\begin{equation}\label{eq:o=o=o}
    o_1 = o_{l+1} = o,\nonumber 
\end{equation}
i.e., any meta-path is selected if its length is within a specified range and its first and last nodes are of the same type. This condition ensures homogeneity in the resulting meta-path graphs for each node type \( o \in O_V \), denoted as \( P_o^1, P_o^2, \dots, P_o^{k_o} \). The process is shown on the left side of Figure\ \ref{fig: framework}(a).

Empirically, the \underline{frequency focuses} of these diverse meta-path graphs are distinct from each other, i.e., their spectral energy distributions exhibit different patterns.
This observation motivates us to assign dedicated Chi-Square filter to each meta-path graph, ensuring that the frequency focus of the filter aligns with that of the graph. This approach has proven effective in homogeneous GAD scenarios \cite{GHRN, AMNet, BWGNN}. Based on this, we detail this process in three steps. First, we derive the frequency focus of both meta-path graphs (step 1) and Chi-Square filters (step 2). Then, we present the filter assignment strategy to generate final representations (step 3).

\header{\bf Step 1: Meta-Path Graph Frequency Focus Estimation.} 
Although the spectral energy \cite{BWGNN, GHRN} has comprehensively characterized each meta-path graph from the spectral domain, it involves eigen-related computation, which is highly time-consuming. Thus, we do not calculate the energy distribution for all meta-path graphs. Instead, we use $S_{high}$ as mentioned in Section \ref{subsec: S_high intro}, which has been shown to reflect the frequency characteristics of the graph effectively \cite{BWGNN, GHRN}, to select a small number of representative meta-path graphs for the aforementioned computation. Particularly, we divide all the valid meta-path graphs into three divisions of equal size by ranking on $S_{high}$, identifying them as low, mid, and high divisions, respectively. Then, based on this division and the proven close connection between $S_{high}$ and spectral energy \cite{BWGNN, GHRN}, the meta-path graph with median $S_{high}$ in each division is regarded as the representative of this division, i.e. its spectral energy is representative of other meta-path graphs in the division. They can be denoted as $G_{P_o}^{(b)}$ for node type $o\in O_V$ and $b \in \{low, mid, high\}$. Thereby, we then fine-grainedly examine their frequency focuses. Denote the eigenvalues (sorted in ascending order), eigenvectors, Laplacian matrix of representative meta-path graph $G_{P_o}^{(b)}$ and the $k$-th column of attribute matrix as $\{\lambda_j^{(b)}\}_{j=1}^{|V_o|}~$, $\vect{U}_{[:,j]}^{(b)}~$, $\vect{L}_o^{(b)}$ and $\vect{x}_o^k$, respectively, where $k=1,\cdots,d_o$. The Fourier coefficients are expressed as: 
\begin{align}\label{eq:fouriercoef}
    \tilde{\vect{x}}^{(b)}_{o,j} = \sum_{k=1}^{|d_o|} \left(\vect{U}_{[:,j]}^{(b)}\right)^T \vect{x}_o^{k}, \quad j = 1, 2, \cdots, |V_o|,
\end{align}
which capture the spectral energy of the graph \( G_{P_o}^{(b)} \). The energy associated with each eigenvalue \( \lambda_j^{(b)} \) is defined as
\(
w^{(b)}_j = \left(\tilde{\vect{x}}^{(b)}_{o,j}\right)^2.
\)
Rather than considering the contribution of each eigenvalue individually, we exploit the excellent band-pass property of the Chi-Square filter by analyzing the focus on a band-by-band basis. To this end, we partition the full frequency band \([0,2]\) into \(K\) sequential, disjoint bands. Formally, for each band \( k \), we define:
\begin{align}\label{eq:bandk}
    \text{Band}_{k} &= \text{List}\left[\lambda_j^{(b)}\right]_{j=k}^{k+\frac{|V_o|}{K}-1}, \quad
    w_{\text{Band}_k}^{(b)} = \sum_{j=k}^{k+\frac{|V_o|}{K}-1} w_j^{(b)},
\end{align}
with \( k \in \{1,\, 1+\frac{|V_o|}{K},\, 1+2\frac{|V_o|}{K},\, \dots,\, |V_o|-\frac{|V_o|}{K}+1\} \). The band with the maximum cumulative weight is then identified as
\begin{equation}\label{eq:bandmaxb}
    \text{Band}_{\max}^{(b)} = \text{REP}\Bigl(\arg\max_{\text{Band}_k} w_{\text{Band}_k}^{(b)}\Bigr) \in \mathcal{R}^1,
\end{equation}
where \( \text{REP} \) denotes a representative operation on \( \text{Band}_k \) (e.g., computing the median or mean). Consequently, the spectral focus for all meta-path graphs in division \( b \) is defined to be \( \text{Band}_{\max}^{(b)} \).

\header{\bf Step 2: Obtaining Frequency Focus for Chi-Square Filters.} 
Unlike the complex band analysis used for meta-path graphs, the frequency focus of each Chi-Square filter \(f_i\) (\(i = 1, 2, \dots\)) can be defined more intuitively. Specifically, we identify the most prominent frequency as
\(
f_i^{-1}\bigl(f_i^{\max}\bigr) \in \mathbb{R}^1,
\)
where \(f_i^{\max}\) is the maximum value of \(f_i(w)\) (defined in Equation \ref{eq: f_i}), and \(f_i^{-1}\) is its inverse function.

\header{\bf Step 3: Assigning Filter and Semantic Representation Learning.} 
Since now both the frequency focuses of meta-path graphs and Chi-Square filters are available, it is now feasible to obtain the Chi-Square filter $f_o^{(b)}$ that mostly aligns with the frequency focus $\text{Band}_{\max}^{(b)}$ of $G_{P_o}^{(b)}$ by
\begin{equation}\label{eq:f0b=argmin}
  f_o^{(b)} = \arg\min_{f_i} |f_i^{-1}(f_i^{\max})-\text{Band}_{\max}^{(b)}|
\end{equation}
and subsequently learn the node representation across diverse meta-path graphs with assigned Chi-Square filters. Prior to the final representation learning, the further process for meta-path graphs and assigned filters are conducted, respectively. For meta-path graph side, weight $w_{P_o^i}^S$ is assigned to the graph shift operator $\vect{S_{P_o^i}}$ of each meta-path graph $G_{P_o^i}$ to learn the distinct importance of diverse meta-paths; for Chi-Square filter side, based on the additivity of the Chi-Square distribution, we propose a novel fusion method Chi-Square filter, which is formally defined as: 
\begin{equation}\label{eq:fpoi}
    f_{P_o^i} = \Big(\prod_{b \in \mathcal{B}}\hspace{-0.5mm}\ast\Big)~w_{P_o^i}^{(b)}f_o^{(b)}, \nonumber
\end{equation}
where $f_{P_o^i}$ is the filter assigned for each specific meta-path graph $G_{P_o^i}$, $\prod\ast$ is continuous convolution between filters, $\mathcal{B}=\{low, mid,\\ high\}$, and $w_{P_o^i}^{(b)}$ is a division-wise weight defined as 
\begin{equation}\label{eq:wpoib}
    w_{P_o^i}^{(b)} = \begin{cases}
        1,\quad \text{if $G_{P_o^i}$ is ranked as a $(b)$-band-focused graph}\\
        w_d,\quad \text{if $G_{P_o^i}$ is not ranked as a $(b)$-band-focused graph}
    \end{cases} \nonumber
\end{equation}
to retain the focus of the convolved filter $f_{P_o^i}$ in the bands highlighted by the spectral energy of $G_{P_o^i}$ while introducing neighboring bands as supplements. $w_d \ll 1$ serves as a hyper-parameter. To make such a form of convolution clearer.

In summary, following the definition of multi-graph filter in Equation \ref{eq: H}, the Multi-Graph Chi-Square Filter $H_o$ formally processes the input representation $\vect{X}_o$ as:
\begin{align}\label{eq:xso=H}
    \vect{X}_o^{s} &= \text{Sum}\Big(\Big[ w_{P_o^1}^S, \cdots, w_{P_o^{k_o}}^S \Big] \odot \Big[ \vect{S}_{P_o^1},\cdots, \vect{S}_{P_o^{k_o}}\Big] \odot \Big[ f_{P_o^1},\cdots, f_{P_o^{k_o}}\Big] \Big) \nonumber \\
    &= \sum_{i=1}^{k_o} f_{P_o^i}\Big(w_{P_o^i}^S \vect{S}_{P_o^i}\Big)\vect{X}_o = H_o(\vect{S}_{P_o^1},\cdots,\vect{S}_{P_o^{k_o}})\vect{X_o} 
\end{align}    
for any $o\in O_V$, where $\sigma$ is an activation function. $\vect{X}_o^{s}$ serves as \textbf{semantic representation} of nodes of type $o$, where semantics across various meta-paths and diverse frequency bands are integrated. In this process, the additivity property of the Chi-Square plays an essential role in two aspects:
\begin{itemize}[topsep=0.5mm, partopsep=0pt, itemsep=0pt, leftmargin=10pt]
    \item The additivity of Chi-Square distribution ensures that each monomial $f_{P_o^i}$ within the convolved filter $H_o$ still aligns with the definition of graph wavelet, retaining excellent properties detailed in Section \ref{sec: wavelet properties}. Thus, the representation $\vect{X}_o^S$ is a linear combination of the learned outputs from multiple Chi-Square filters.
    \item The additivity of Chi-Square promotes collaborative representation learning across diverse meta-paths in an effective way. Specifically, it not only ensures the graph with signals are processed by the meticulously assigned and most aligned filter as mentioned but also incorporates additional filters naturally derived from convolutional fusion to enhance broader frequency-band coverage. 
\end{itemize}

\subsection{Interactive Meta-Graph Convolution}
\label{sec: method II interactive xxx}
Processing each node type simultaneously leaves dimensional disparities unresolved—a major challenge in heterogeneous settings. Moreover, retaining high-frequency information is crucial for GAD tasks, as it helps counteract over-smoothing \cite{BWGNN, GHRN, AMNet}. Yet, the ability of alignment methods to preserve high-frequency content during dimensional alignment in HINs remains largely unexplored.

In this study, we propose to use the linear layer to align feature dimensions. Despite its simplicity, our provided theoretical analysis in this section and empirical results in Appendix \ref{app: experiments for comparison of alignment method} have jointly proven that the linear layer is a nice choice as the alignment component of the GAD model in heterogeneous scenario by effectively preserving high-frequency contents. Next, we first introduce an intuitive revelation of the linear layers' functionality.

\header
\textbf{An Intuitive Revelation of the Functionality.} For nodes $V_o$ of any type $o\in O_V$, the semantic representation matrix learned from Equation \ref{eq:xso=H} is comprised of $d_o^{s}$ signals as column vectors, which is formally expressed as:
\begin{equation}\label{eq:x_so=xxxx}
\vect{X}_o^{s} = \vect{x}_o^{s1} \oplus \vect{x}_o^{s2} \oplus \cdots \oplus \vect{x}_o^{sd_o^{s}}  \in \mathcal{R}^{|V_o|\times d_o^{s}}.     \nonumber
\end{equation}
Given a node-type-wise linear layer $\vect{W}_o=(w_{ij})_{d_o^{s} \times d_a}\in \mathcal{R}^{d_o^{s} \times d_a}$ where $d_a$ represents the aligned dimension, the aligned representation is computed by:
\begin{equation}\label{eq: X=XW}
    \vect{X}_o^a = \vect{X}_o^{s}\vect{W}_o =  \vect{x}_o^{a,1} \oplus \vect{x}_o^{a,2} \oplus \cdots \oplus \vect{x}_o^{a,d_o} \in \mathcal{R}^{|V_o|\times d_a}.
\end{equation}
Equation \ref{eq: X=XW} reveals the derivation of each aligned representation $x_o^{a,j}$ is a linear summation of the raw graph signals as:
\begin{equation}\label{}
    \vect{x}_o^{a,j}=\sum_{i=1}^{d_0}w_{ij} \vect{x}_o^{si},~j=1,2,\cdots,d_a, \nonumber
\end{equation}
which indicates the linear layer $\vect{W}_o$ learns each new graph signal $\vect{x}_o^{a,j}, ~j=1,\cdots,d_a$ by automatically adjusting the relationships between the raw graph signals $\{\vect{x}_o^{si}\}_{i=1}^{d_o}$ with learnable linear weights.

\header
\textbf{Theoretical Justification of Linear Summation.} The vital importance of counteraction against over-smoothness and retaining high-frequency to GAD task have been deeply investigated in the latest researches \cite{BWGNN, GHRN, AMNet}; meanwhile, $S_{high}$ serves as a handy depiction for the smoothness and high-frequency content of any graph signal from the spatial and spectral domains, respectively \cite{BWGNN, GHRN}. Based on these two findings, we then prove the linear layer's excellent capability of high-frequency retention.

\begin{theorem}\label{thm: 1}
For a given collection of $k$ graph signals $x_1, \cdots, x_{k}$ defined on a graph with Laplacian matrix $L$. In theory, there exists a linear alignment layer such that the aligned node representation $x_a$ perfectly keeps the $S_{high}$. Formally, $S_{high}(x_a,L)=\max\big\{S_{high}(x_i,L)\big\}_{i=1}^{k}$.
\end{theorem}

Proof of Theorem \ref{thm: 1} is presented in Appendix \ref{sec: proof for thm1}. Theorem \ref{thm: 1} illuminates the accessibility of optimal alignment with linear layers when the preservation of $S_{high}$ is taken as criterion. Notably, the fact that the linear transformations represented by $\vect{W}_o$ are not constant is not a detriment to performance in both retention and GAD tasks; on the contrary, this flexibility is shown to be beneficial in experimental results of Appendix \ref{app: experiments for comparison of alignment method}, probably because it enables $\vect{W}_o$ to automatically learn the linear mapping that is most effective for anomaly detection; in addition, other alignment methods with similar linear shapes, e.g. PCA, AutoEncoder also show significant performance improvements over nonlinear methods in both high-frequency retention and GAD tasks in Appendix \ref{app: experiments for comparison of alignment method}, which further verifies our theoretical findings. These subsequent theoretical and empirical results motivate us to propose to align the representations of diverse nodes as:
\begin{equation}\label{eq: aligned node repres}
    \vect{X}_o^a = \vect{X}_o^{s}\vect{W}_o, \quad o=1,2,\cdots,|O_V|,
\end{equation}
where $\vect{W}_o \in \mathcal{R}^{d_o^{s}\times d_a}$ is a node-type-wise linear layer and $\vect{X}_o^a$ is the aligned representation of type $o$ nodes. Then, employing Method 1 as detailed in Section \ref{method 1}, we derive the transformed homogeneous graph $G_M^{'}$ comprising all types of nodes, and thence learn the final node representations formally as:
\begin{equation}\label{eq: CHIGNN}
    \vect{X}_o^{'} = \sum_{f\in \mathcal{F}_{M^{'}}}f(\vect{S}_{M^{'}})\sigma(\vect{X}_o^{a}), \quad o=1,2,\cdots,|O_V|,
\end{equation}
where $\mathcal{F}_{M^{'}}$ is a specified set of Chi-Square filters covering diverse frequency bands and $\vect{S}_{M^{'}}$ is the graph shift operator of $G_{M^{'}}$.

\subsection{Contribution-Informed Cross-Entropy Loss}\label{sec: cc_loss}
Prior works \cite{RQGNN, GHRN} have highlighted the significant challenges posed by the class imbalance inherent in anomaly graphs within GAD. This motivates us to develop this module to enhance the effective learning of anomalies.Specifically, in this module, we firstly obtain a homogeneous graph $G_{o_t}$for the target node type $o_t$ using Method 2 delineated in Section \ref{method 1}. For the final representation $\vect{X}_{o_t}^{'} = \vect{x}_{o_{t}}^{'1} \oplus \vect{x}_{o_{t}}^{'2} \oplus \cdots \oplus \vect{x}_{o_{t}}^{'d_{o_{t}}^{'}}  \in R^{|V_{o_{t}}|\times d_{o_{t}}^{'}}$ and the Laplacian matrix $\vect{L}_{o_{t}}$ of $G_{o_t}$, the contribution of each node is defined as:
\begin{equation}\label{}
    c_i = 
    \sum_{j=1}^{d}
    \frac{\vect{x}_{o_t}^{'j,i}(\vect{L}_{o_t}\vect{x}_{o_t}^{'})^{j,i}}{{\vect{x}_{o_t}^{'j}}^{T}\vect{L}_{o_t}\vect{x}_{o_t}^{'j}},
    ~i=1,\cdots,|V_{o_t}|, \nonumber
\end{equation}
where $\vect{x}_{o_t}^{'j,i},(\vect{L}_{o_t}\vect{x}_{o_t}^{'})^{j,i}\in\mathcal{R}^1$ are the $i$-th element of column vectors $\vect{x}_{o_t}^{'j}$ and $ \vect{L}_{o_t}\vect{x}_{o_t}^{'j}$, respectively. With a focus on the distribution of contributions, the following empirical observations can be revealed:
\begin{itemize}[topsep=0.5mm, partopsep=0pt, itemsep=0pt, leftmargin=10pt]
    \item In most cases, the average contribution of anomalies is significantly larger than that of benign nodes within the train set;
    \item In most cases, the average contribution of recognized anomalies (i.e. true positive/TP) is significantly larger than that of unrecognized anomalies (i.e. false negative/FN) within train set,
\end{itemize}
which jointly indicate that the benign and FN anomalies within the train set exhibit similar overall deficiencies in terms of contributions to $S_{high}$. 
 Such similarity is not surprising, since the representations of FN anomalies are generally more challenging to distinguish from benign nodes than those of TP anomalies.

\begin{table*}[htbp!]
\caption{Experimental results of performance comparison on heterogeneous datasets (ACM, R-I, R-II).}
\vspace{-4mm}
\begin{adjustbox}{max width=\textwidth} 
\label{tab: het - performance evaluation}
\begin{tabular}{c|cccc|cccc|cccc}
\hline
Datasets          & \multicolumn{4}{c|}{ACM}                                              & \multicolumn{4}{c|}{R-I}                                              & \multicolumn{4}{c}{R-II}                                              \\
Metrics           & AUROC           & AUPRC           & F1-macro        & Recall          & AUROC           & AUPRC           & F1-macro        & Recall          & AUROC           & AUPRC           & F1-macro        & Recall          \\ \hline
Operational-Model & -               & -               & -               & -               & 0.8200          & 0.0473          & 0.5528          & 0.1105          & 0.8120          & 0.0482          & 0.5520          & 0.1378          \\
PSHGCN            & 0.9564          & 0.6226          & 0.7992          & 0.7305          & 0.8650          & 0.2264          & 0.6425          & 0.3316          & 0.9122          & 0.4355          & 0.7148          & 0.4551          \\
BWGNN             & 0.9030          & 0.8452          & 0.4015          & 0.7444          & 0.8444          & 0.2156          & 0.6491          & 0.3263          & 0.9133          & 0.2231          & 0.6197          & 0.2231          \\
XGBGraph          & 0.9461          & 0.9068          & 0.8647          & 0.8271          & 0.8795          & 0.2960          & 0.6812          & 0.3632          & 0.9155          & 0.5696          & 0.8014          & 0.5992          \\
ConsisGAD         & 0.8773          & 0.7462          & 0.8066          & 0.7461          & 0.8724          & 0.2944          & 0.6716          & 0.3747          & 0.9014          & 0.4500          & 0.7284          & 0.4696          \\
\textbf{ChiGAD}   & \textbf{0.9702} & \textbf{0.9518} & \textbf{0.9195} & \textbf{0.8947} & \textbf{0.9472} & \textbf{0.3887} & \textbf{0.7272} & \textbf{0.4579} & \textbf{0.9322} & \textbf{0.5854} & \textbf{0.8103} & \textbf{0.6133} \\ \hline
Improvement       & 2.55\%          & 4.96\%          & 6.34\%          & 8.17\%          & 7.70\%          & 31.34\%         & 6.75\%          & 26.09\%         & 1.82\%          & 2.77\%          & 1.11\%          & 2.37\%          \\ \hline
\end{tabular}
\end{adjustbox}
\vspace{-1mm}
\end{table*}

Building upon the crucial findings and analyses, we propose the Contribution-Informed Cross-Entropy Loss, formally defined as:
\begin{equation*}\label{}
    \mathcal{L}_{CC} = -\frac{1}{N}\sum_{i=1}^{N}w_i\Big[y_i\log(\hat{p}_i) + (1-y_i)\log(1-\hat{p}_i)\Big], \nonumber
\end{equation*}
where for any $i$-th sample in the train set, $y_i$ is the true label,
$$
    \hat{p}_i = \text{softmax}(\vect{X}_{o_t}^{'})_i
$$
is the predicted anomaly probability, and 
\begin{equation*}\label{}
    w_i = \begin{cases}
        1, &\text{ if $i$-th sample } \in V^{bn},\\
        \frac{c_{\max}-c_i}{c_{\max}-c_{\min}}(H-L)+L, &\text{ if $i$-th sample } \in V^{an}, \nonumber
    \end{cases}
\end{equation*}
is the contribution-informed weight. $~H,~L$ are hyper-parameters to regulate the range of weights, and $c_{\min},~c_{\max}$ are computed by:
\begin{equation*}\label{}
    c_{\min} = \min\{c_i\}_{i=1}^{N}, ~c_{\max} = \max\{c_i\}_{i=1}^{N}. \nonumber
\end{equation*}
Notably, we impose an essential constraint that $H \geq L \geq 1$,
which ensures that the model prioritizes challenging anomalies (FN) and gives greater attention to anomalous samples, which are of scarcity, compared to benign nodes. As the weights assigned to anomalous samples are ensured to be greater, this constraint helps mitigate the inherent class-imbalance problem in anomaly detection tasks.

\section{Experiments}
\label{sec:experiments}

In this section, we conduct holistic experiments to evaluate the capability of ChiGAD in heterogeneous graph anomaly detection.

\begin{table*}[htbp]
\caption{Experimental results of performance comparison on homogeneous datasets.}
\label{tab:homo-performance-evaluation-1}
\vspace{-4mm}
\begin{adjustbox}{max width=\textwidth}
\begin{tabular}{cc|cccccccccccc}
\hline 
Datasets                   & Metrics  & GAT    & GIN     & PCGNN  & BernNet & PSHGCN & AMNet  & BWGNN  & GAGA   & GHRN   & XGBGraph & ConsisGAD & \textbf{ChiGNN} \\ \hline
\multirow{4}{*}{Reddit}    & AUROC    & 0.6415 & 0.6182  & 0.6541 & 0.6983  & 0.6744 & 0.6960 & 0.7082 & 0.5352 & 0.6102 & 0.6474   & 0.5859    & \textbf{0.7288} \\ 
                           & AUPRC    & 0.0720 & 0.0641  & 0.0773 & 0.0782  & 0.0625 & 0.0787 & 0.0832 & 0.0359 & 0.0466 & 0.0529   & 0.0607    & \textbf{0.0953} \\
                           & F1-macro & 0.4915 & 0.4915  & 0.4915 & 0.5406  & 0.4915 & 0.4915 & 0.5129 & 0.4915 & 0.5149 & 0.5286   & 0.4915    & \textbf{0.5596} \\
                           & Recall   & 0.1088 & 0.1088  & 0.1497 & 0.1088  & 0.0952 & 0.1156 & 0.1156 & 0.0204 & 0.0544 & 0.0612   & 0         & \textbf{0.1565} \\ 
                           \hline
\multirow{4}{*}{Tolokers}  & AUROC    & 0.7875 & 0.7457  & 0.7663 & 0.7751  & 0.7784 & 0.7516 & 0.8041 & 0.5001 & 0.8008 & 0.8285   & 0.7972    & \textbf{0.8349} \\
                           & AUPRC    & 0.4525 & 0.4036  & 0.4485 & 0.4369  & 0.4494 & 0.4074 & 0.4958 & 0.2184 & 0.4745 & 0.5392   & 0.4691    & \textbf{0.5413} \\
                           & F1-macro & 0.6356 & 0.6087  & 0.6047 & 0.6152  & 0.5568 & 0.6200 & 0.6574 & 0.4387 & 0.6484 & 0.6688   & 0.6741    & \textbf{0.7137} \\
                           & Recall   & 0.4424 & 0.3941  & 0.4315 & 0.4470  & 0.4517 & 0.4252 & 0.5031 & 0.2150 & 0.4657 & 0.5343   & 0.5252    & \textbf{0.5514} \\ 
                           \hline
\multirow{4}{*}{Amazon}    & AUROC    & 0.9713 & 0.9563  & 0.9801 & 0.9579  & 0.9726 & 0.9692 & 0.9827 & 0.9731 & 0.9829 & 0.9874   & 0.9823    & \textbf{0.9878} \\
                           & AUPRC    & 0.8794 & 0.8461  & 0.8933 & 0.8489  & 0.8855 & 0.8836 & 0.9148 & 0.8807 & 0.8952 & 0.9333   & 0.9273    & 0.9160          \\
                           & F1-macro & 0.9034 & 0.2608  & 0.8997 & 0.8941  & 0.9072 & 0.8956 & 0.9034 & 0.9155 & 0.9039 & 0.9276   & 0.9265    & \textbf{0.9300} \\
                           & Recall   & 0.8261 & 0.8098  & 0.8533 & 0.8261  & 0.8533 & 0.8315 & 0.8587 & 0.8376 & 0.8533 & 0.8533   & 0.8061    & \textbf{0.8859} \\ 
                           \hline
\multirow{4}{*}{T-Finance} & AUROC    & 0.9575 & 0.9271  & 0.9403 & 0.9667  & 0.9735 & 0.9638 & 0.9693 & 0.9439 & 0.9646 & 0.9715   & 0.9736    & \textbf{0.9749} \\
                           & AUPRC    & 0.8272 & 0.7835  & 0.8331 & 0.8917  & 0.8961 & 0.8887 & 0.8938 & 0.8079 & 0.8760 & 0.9012   & 0.8937    & \textbf{0.9029} \\
                           & F1-macro & 0.7296 & 0.8254  & 0.7949 & 0.7221  & 0.9141 & 0.8273 & 0.6505 & 0.8041 & 0.6484 & 0.9308   & 0.9191    & \textbf{0.9319} \\
                           & Recall   & 0.7975 & 0.7337  & 0.7906 & 0.8363  & 0.8377 & 0.8405 & 0.8419 & 0.7916 & 0.8197 & 0.8502   & 0.8278    & \textbf{0.8544} \\ 
                           \hline
\multirow{4}{*}{Questions} & AUROC    & 0.6830 & 0.6808  & 0.6766 & 0.6858  & 0.6935 & 0.6605 & 0.7087 & 0.5003 & 0.7216 & 0.7102   & 0.7316    & \textbf{0.7373} \\
                           & AUPRC    & 0.1551 & 0.1368  & 0.1559 & 0.1725  & 0.1394 & 0.1563 & 0.1857 & 0.0299 & 0.1831 & 0.1819   & 0.1797    & \textbf{0.1930} \\
                           & F1-macro & 0.5726 & 0.5709  & 0.5799 & 0.5778  & 0.5572 & 0.5780 & 0.5878 & 0.4924 & 0.5857 & 0.5879   & 0.5896    & \textbf{0.6057} \\
                           & Recall   & 0.1726 & 0.1836  & 0.1808 & 0.1973  & 0.1753 & 0.1753 & 0.2164 & 0.0301 & 0.2219 & 0.2055   & 0.1340    & \textbf{0.2329} \\ 
                           \hline
\multirow{4}{*}{DGraph-Fin} & AUROC    & 0.7553 & 0.7416  & 0.7276 & 0.7358  & 0.7010 & 0.7299 & 0.7630 & 0.5000 & 0.7613 & 0.7583   & 0.7040    & \textbf{0.7640} \\
                           & AUPRC    & 0.0385 & 0.0347  & 0.0342 & 0.0327  & 0.0335 & 0.0281 & 0.0397 & 0.0127 & 0.0380 & 0.0379   & 0.0080    & \textbf{0.0403} \\
                           & F1-macro & 0.5317 & 0.5150  & 0.5072 & 0.5199  & 0.4968 & 0.5177 & 0.5309 & 0.4968 & 0.5323 & 0.5259   & 0.4989    & \textbf{0.5344} \\
                           & Recall   & 0.0714 & 0. 0632 & 0.0666 & 0.0555  & 0.0469 & 0.0421 & 0.0757 & 0.0125 & 0.0696 & 0.0696   & 0         & \textbf{0.0800} \\
                           \hline
\multirow{4}{*}{T-Social}  & AUROC    & 0.9033 & 0.9409  & 0.8769 & 0.9373  & 0.9184 & 0.9250 & 0.9688 & 0.9696 & 0.9712 & 0.9976   & 0.9636    & \textbf{0.9980} \\
                           & AUPRC    & 0.3207 & 0.6079  & 0.8029 & 0.4430  & 0.7111 & 0.3770 & 0.7893 & 0.8345 & 0.8678 & 0.9734   & 0.7213    & \textbf{0.9833} \\
                           & F1-macro & 0.5740 & 0.5462  & 0.4801 & 0.4341  & 0.8241 & 0.4879 & 0.6304 & 0.7319 & 0.6455 & 0.9256   & 0.8224    & \textbf{0.9760} \\
                           & Recall   & 0.4207 & 0.6447  & 0.7353 & 0.4823  & 0.7034 & 0.4321 & 0.7578 & 0.7393 & 0.8233 & 0.9353   & 0.6089    & \textbf{0.9535} \\ 
                           \hline
\end{tabular}
\end{adjustbox}
\end{table*}

\subsection{Experimental Setup}

\header
\textbf{Datasets.}
We assess ChiGAD using one public and two industrial datasets, and further evaluate ChiGNN, a ChiGAD-derived homogeneous graph anomaly detector, on seven public datasets:
\begin{itemize}[topsep=0.5mm, partopsep=0pt, itemsep=0pt, leftmargin=10pt]
    \item \textbf{Heterogeneous graph for ChiGAD}: Public ACM dataset and two real-world datasets R-I and R-II provided by WeChat Pay, Tencent, which pertain to financial activities in two different regions.
    \item \textbf{Homogeneous graph for ChiGNN}: Amazon, T-Finance, T-Social, Reddit, Tolokers, Questions, and DGraph-Fin.
\end{itemize}
The summaries of datasets are detailed in Appendix \ref{app: profiles of datasets}. It is worth noting that ACM is not a natural dataset for GAD tasks. However, the node of type $o_0$ has 3 classes, which allows us to fix the target node type $o_t=o_0$ and randomly select one of the classes as the anomalous class and the other two as the benign class.

\header
\textbf{Baselines and Implementation Details.}
We introduce a diverse range of cutting-edge baselines for different types of experiments to compare and evaluate the model performance:
\begin{itemize}[topsep=0.5mm, partopsep=0pt, itemsep=0pt, leftmargin=10pt]
    \item \textbf{Baselines for ChiGAD}: Operational-Model, PSHGCN \cite{PSHGCN}, BWGNN \cite{BWGNN}, XGBGraph \cite{XGBGraph}, and ConsisGAD \cite{ConsisGAD};
    \item \textbf{Baselines for ChiGNN}: GAT \cite{GAT}, GIN \cite{GIN}, PCGNN \cite{PCGNN}, BernNet \cite{BernNet}, PSHGCN \cite{PSHGCN}, AMNet \cite{AMNet}, BWGNN \cite{BWGNN}, GAGA \cite{GAGA}, GHRN \cite{GHRN}, XGBGraph \cite{XGBGraph}, and ConsisGAD \cite{ConsisGAD}.
\end{itemize}
Note that, for fair comparison in all experiments, we obtain the source code of these baselines from GitHub and use the default hyper-parameter setting recommended by their authors. In addition, to ensure compatibility with models not originally designed for heterogeneous graphs, we degenerate heterogeneous graphs, ACM, R-I, and R-II, into homogeneous graphs through two distinct methods, and adopt the \textbf{superior} performance obtained from these two methods for a fair comparison:
\begin{itemize}[topsep=0.5mm, partopsep=0pt, itemsep=0pt, leftmargin=10pt]
    \item \textbf{Method 1:} All the nodes within the meta graph, i.e. the raw heterogeneous graph, are reserved; any edge is reserved if and only if it exists in at least one relation. In addition, the problem of dimensional disparity is addressed initially with linear layer;\label{method 1}
    \item \textbf{Method 2:} We keep only the nodes of target type $o_t$. The inter-connection is formed only when the endpoints are connected in the context of any considered relation. 
\end{itemize}

As for the implementation setup, we employ the fully-supervised settings following the up-to-date benchmark GADBench \cite{XGBGraph}. Detailed information for reproducibility is in Appendix \ref{app: data split} and \ref{app: hyper-parameter setting}.

\header
\textbf{Metrics.}
In this study, \textbf{AUROC, AUPRC, F1-macro}, and \textbf{Recall} are selected for providing reliable and holistic evaluation, since the former two are threshold-free while the rest are threshold-based.

\subsection{Performance Comparison}\label{subsec: performance comparison}
Table \ref{tab: het - performance evaluation} shows the performance of ChiGAD against diverse baselines across 3 heterogeneous graphs. We highlight the best performance among competitors in bold text. 

Firstly, ChiGAD significantly surpasses all the state-of-the-art methods in all heterogeneous datasets across all metrics, demonstrating its superiority in heterogeneous graph anomaly detection.

Secondly, the explicit capture of rich semantics across diverse meta-path graphs has been demonstrated to be crucial for heterogeneous graph anomaly detection. On one hand, ChiGAD and PSHGCN, as multi-graph filters, both explicitly introduce various graph shift operators to capture semantics based on various meta-paths, and both show excellent GAD capabilities on the three datasets (it is worth noting that PSHGCN is not designed for GAD tasks); on the other hand, in contrast, despite being state-of-the-art methods in the field of general homogeneous GAD, ConsisGAD and BWGNN perform worse than PSHGCN due to their lack of multi-graph processing capabilities.

Then, we further compare ChiGAD with the two models designed for heterogeneous graph learning, i.e. Operational-Model and PSHGCN. (1) Operational-Model is a model actually applied in the business of company R. It is based on GCN and learns features on multiple manually selected important meta-path graphs. Due to privacy reasons, Operational-Model cannot be obtained for offline deployment on the ACM dataset. Empirically, it shows the worst performance on the business datasets R-I and R-II. This can be attributed to the low-pass property of GCN, which causes a large amount of information loss when facing input signals containing diverse frequency band information, which is crucial for GAD \cite{BWGNN}. (2) Although PSHGCN is also a multi-graph filter, it is significantly inferior to ChiGAD across all the datasets and all the metrics. This proves that our proposed key components of ChiGAD designed for GAD help detect anomalies in heterogeneous graphs effectively.

ChiGAD significantly outperforms ConsisGAD, BWGNN, and XGBGraph in heterogeneous graph anomaly detection, as evidenced by superior performance across AUROC, AUPRC, F1-macro, and Recall on ACM, R-I, and R-II datasets. While ConsisGAD addresses limited supervision through consistency training, it lacks explicit handling of heterogeneity. BWGNN, designed for homogeneous graphs, struggles with the complexity of heterogeneous networks, and XGBGraph, despite its neighbor aggregation strength, fails to model fine-grained relational properties in heterogeneous graphs. In contrast, ChiGAD leverages multi-graph Chi-Square filters to capture cross-meta-path anomalies, preserves high-frequency content via linear alignment, and mitigates class imbalance with contribution-informed loss, showcasing the inherent advantages of multi-graph filters in heterogeneous anomaly detection.

\subsection{Superiority of Chi-Square Filter}\label{}
In this study, the Chi-Square filter serves as the foundation for the multi-graph filter. Furthermore, as shown in Equation \ref{eq: CHIGNN}, in scenarios where dimensional disparities are not a concern, the Chi-Square convolutional layers, namely \textbf{ChiGNN}, can be separately compatible for homogeneous graph anomaly detection tasks. To investigate this, we examined the performance of as many as seven public datasets, with the experimental results presented in Table \ref{tab:homo-performance-evaluation-1}. It is evident that the ChiGNN outperforms all baselines in GAD tasks. This can be attributed to the excellent properties with the Chi-Square filter, e.g., both the spatial and spectral locality, as well as the wide coverage on diverse frequency bands. For reproducibility, the hyper-parameter settings are listed in Appendix \ref{app: hyper-parameter setting}. 

\subsection{Resisting Over-Smoothness for GAD: Comparison of Alignment Methods}
\label{app: experiments for comparison of alignment method}

\begin{figure}[!h]
    \centering
     \includegraphics[width=0.45\textwidth]{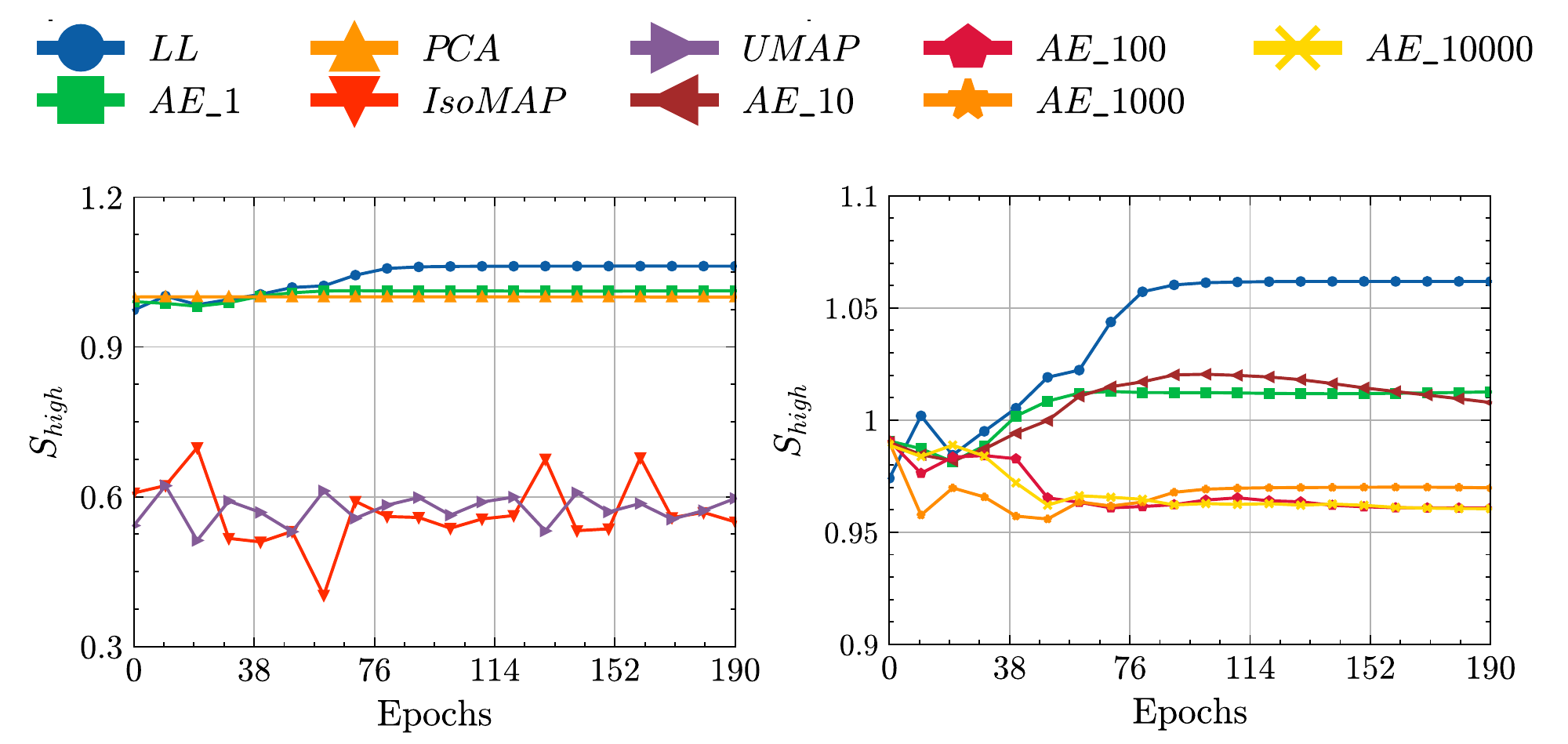}
\vspace{-3mm}
\caption{Comparison of results of frequency retention task across diverse alignment methods on ACM.}
\vspace{-3mm}
\label{fig: comparison of alignment method on frequency retention}
\end{figure}

This section aims at comparing the performance of diverse alignment methods on both the frequency retention task and the GAD task, which further helps reveal the relationship between resisting over-smoothness and anomaly detection. To this end, we first compare the performance of linear layer (LL), which is employed as dimension alignment component of ChiGAD as detailed in Section \ref{sec: method II interactive xxx} with diverse representative alignment methods, i.e. Autoencoder (AE), Principle Component Analysis (PCA), Uniform Manifold Approximation and Projection(UMAP), and Isometric Mapping (IsoMAP) on the frequency retention task. Based on this, we further compare the performance of ChiGAD and four variants with AE, PCA, UMAP, and IsoMAP as the alignment components on GAD tasks. It is worth noting that 
we take $S_{high}$ as the metric to quantify both the smoothness content and high-frequency content retained (the frequency energy distribution of graph signal in the spectral domain, which is evaluated by $S_{high}$, is directly and closely related to its smoothness as node representation in the spatial domain \cite{BWGNN, GHRN}. Specifically, higher $S_{high}$ indicates a lower degree of smoothness and more high-frequency content is preserved from the spatial and spectral domains, respectively, and vice versa. \cite{BWGNN, GHRN}); for the GAD task, AUROC, AUPRC, F1-macro and Recall are introduced. Experimental results are shown in Figure \ref{fig: comparison of alignment method on frequency retention}
and Table \ref{tab: comparison of alignment method on GAD}.

\begin{table}[t]
\caption{Comparison of results of GAD task across diverse alignment methods on ACM.}
\vspace{-4mm}
\label{tab: comparison of alignment method on GAD}
\begin{tabular}{@{}ccccc@{}}
\toprule
Metric & AUROC & AUPRC & F1-macro & Recall \\ \midrule
\textbf{LL (employed)} & \textbf{0.9702} & \textbf{0.9518} & \textbf{0.9195} & \textbf{0.8947} \\ \midrule
$PCA$ & 0.6587 & 0.5035 & 0.5975 & 0.4687 \\
$UMAP$ & 0.5746 & 0.3852 & 0.5393 & 0.3885 \\
$IsoMAP$ & 0.5714 & 0.3854 & 0.4417 & 0.3784 \\ 
$AE$ & 0.9597 & 0.9088 & 0.9157 & 0.8847 \\
$AE\_10$ & 0.9483 & 0.8928 & 0.8798 & 0.8296 \\
$AE\_100$ & 0.9317 & 0.8782 & 0.8511 & 0.7945 \\
$AE\_1000$ & 0.9072 & 0.8517 & 0.8193 & 0.7845 \\
$AE\_10000$ & 0.8450 & 0.7510 & 0.7536 & 0.6667 \\
\bottomrule
\end{tabular}
\vspace{-2mm}
\end{table}

Figure \ref{fig: comparison of alignment method on frequency retention} shows the $S_{high}$ of aligned nodal representation $\vect{X}_o^a$, which is defined in Equation \ref{eq: aligned node repres}, obtained in the training process. Notably, considering both the Laplacian matrices of any mentioned graph and raw nodal features $\vect{X}_o$ are constant (since the former corresponds to structures of graphs, which are consistent), the initial high-frequency content evaluated by $S_{high}$ is constant. Thus, a higher $S_{high}$ of aligned representation indicates a better performance of any specific alignment method in frequency retention task. Based on this finding, subsequent observations are obtained:
\begin{itemize}[topsep=0.5mm, partopsep=0pt, itemsep=0pt, leftmargin=10pt]
\item Linear layer achieves the optimal performance in frequency retention; furthermore, the performance of linear-related methods (LL, PCA, AE) significantly surpasses that of nonlinear methods (UMAP, IsoMAP). This empirical results highly align with our theoretical analysis in Section \ref{sec: method II interactive xxx}.

\item Although AE is not fully linear due to activation functions, it resembles a stack of linear layers, explaining its comparable performance to LL and PCA.

\item Among linear-related methods, LL is significantly superior to PCA and AE due to its flexibility in prioritizing high-frequency retention. Conversely, the inherent functionalities of PCA (variance maximization) and AE (reconstruction error minimization), despite also being linear-related, may conflict with this task.
\item To further explore the above point, multiple variants of AE are introduced, where the reconstruction errors are amplified with weights from 10 to 10,000. As we can see, the variants with larger weights perform worse in the retention task, which indicates the task of reconstruction contradicts high-frequency retaining.
\end{itemize}

Results of the GAD task in Table \ref{tab: comparison of alignment method on GAD} show similar patterns to those of the frequency retention task, which reveals the crucial connection between GAD and high-frequency retention. Linear methods outperform UMAP and IsoMAP, with LL being the best. AE outperforms PCA, especially with smaller reconstruction weights. This indicates that both PCA's variance maximization and AE's reconstruction conflict with GAD, causing greater information loss with PCA in aligning representation dimensions.

\section{Conclusion}
\label{sec:conclusion}

In this paper, we introduce ChiGAD,  an effective framework for graph anomaly detection on heterogeneous networks using our proposed Chi-Square filter. Extensive experiments show the great efficacy of ChiGAD in heterogeneous GAD. Moreover, our ChiGNN variant surpasses state-of-the-art baselines on homogeneous GAD tasks, further validating the effectiveness of the Chi-Square filter.

\begin{acks}
This work is supported by the RGC GRF grant (No. 14217322) and the Tencent WeChat Rhino-Bird Focused Research Program.
\end{acks}

\bibliographystyle{ACM-Reference-Format}
\balance
\bibliography{CrestGAD}

\appendix

\section{Profiles of Datasets}\label{app: profiles of datasets}

\begin{table*}[htbp]
    \centering
    \caption{Summary of datasets.}
    \vspace{-4mm}
    \label{tab: datasets}
    \small
    \setlength{\tabcolsep}{3pt}
    \begin{tabular}{lccccccc}
        \hline
        Dataset   & \#Node   & \#Edge    & \#Feat.                 & Anomaly \% & Split \% (Train/Val/Test) & Relation & Feat. Type \\ \hline
        ACM (het) & 10,942 & 547,872 & [193,115,250,95] & 32.68/32.07/35.07 & 40/20/40 & Co-authorship & Misc. Info. \\
        R-I (het) & 2,880,722 & 8,781,690 & [14,171,10,13] & 0.35 & 40/20/40 & Account Interaction & Misc. Info. \\
        R-II (het) & 6,168,746 & 15,576,508 & [14,171,10,13] & 0.55 & 40/20/40 & Account Interaction & Misc. Info. \\ \hline
        Reddit & 10,984 & 168,016 & 64 & 3.3 & 40/20/40 & Under Same Post & Text Embedding \\
        Tolokers & 11,758 & 519,000 & 10 & 21.8 & 50/25/45 & Work Collaboration & Misc. Info. \\
        Amazon & 11,944 & 4,398,392 & 25 & 9.5 & 70/10/20 & Review Correlation & Misc. Info. \\
        T-Finance & 39,357 & 21,222,543 & 10 & 4.6 & 40/20/40 & Transaction Record & Misc. Info. \\
        Questions & 48,921 & 153,540 & 301 & 3.0 & 50/25/25 & Question Answering & Text Embedding \\
        DGraph-Fin & 3,700,550 & 4,300,999 & 17 & 1.3 & 70/15/15 & Loan Guarantor & Misc. Info. \\
        T-Social & 5,781,065 & 73,105,508 & 10 & 3.0 & 40/20/40 & Social Friendship & Misc. Info. \\ \hline
    \end{tabular}
\end{table*}

\begin{table*}[t]
\caption{Hyper-parameter setting of ChiGNN for experiments.}
\vspace{-4mm}
  \label{tab: hyper-param on homo G}
  \small
  \setlength{\tabcolsep}{3pt}
  \begin{tabular}{@{}c|ccc|ccccccc@{}}
    \toprule
    Hyper-parameter & ACM & R-I & R-II & Reddit & Tolokers & Amazon & T-Finance & Questions & DGraph-Fin & T-Social \\ \midrule
    learning rate & 0.0001 & 0.006 & 0.006 & 0.0003 & 0.003 & 0.0003 & 0.003 & 0.0001 & 0.01 & 0.01 \\
    dropout rate & 0 & 0 & 0 & 0 & 0.3 & 0.6 & 0.6 & 0.4 & 0.4 & 0 \\
    hidden dimension & 512 & 25 & 48 & 64 & 64 & 64 & 80 & 400 & 32 & 20 \\
    aligned dimension & 512 & 25 & 48 & - & - & - & - & - & - & - \\
    epochs & 200 & 200 & 200 & 1700 & 700 & 1550 & 300 & 150 & 1200 & 4000 \\
    MLP layers & 4 & 2 & 2 & 2 & 2 & 3 & 3 & 4 & 2 & 3 \\
    activation & ReLU & ReLU & ReLU & Tanh & ReLU & LeakyReLU & Tanh & ReLU & ReLU & LeakyReLU \\
    $\mathcal{F}_M$ & - & - & - & 2-11 & 1-15 (odd) & 1-19 (odd) & 1-7 (odd) & 1-15 (odd) & 1-15 (odd) & 1-19 (odd) \\
    H & 2.2 & 2.2 & 2.1 & 1.05 & 4 & 3 & 2.05 & 1.05 & 1.05 & 3.05 \\
    L & 1.9 & 1.8 & 1.9 & 0.95 & 3.5 & 1 & 1.05 & 0.95 & 0.95 & 2.95 \\ \bottomrule
  \end{tabular}
\end{table*}

\begin{table}[t]
\caption{Expectation and most highlighted frequency of \( f_i(w) \).}
\vspace{-4mm}
\begin{tabular}{@{}ccc|ccc@{}} 
\toprule
\( i \) & \( \mathbb{E}[f_i(w)] \) & \( f_i^{-1}(\max) \) & \( i \) & \( \mathbb{E}[f_i(w)] \) & \( f_i^{-1}(\max) \) \\ \midrule
1 & 0.6970 & 0.0000 & 16 & 1.5940 & 1.7638 \\
2 & 0.9603 & 0.6667 & 32 & 1.7126 & 1.8779 \\
4 & 1.2180 & 1.1992 & 64 & 1.7973 & 1.9339 \\
8 & 1.4313 & 1.5556 & 128 & 1.8571 & 1.9600 \\
\bottomrule
\end{tabular}
\label{tab: Combined Expectations and Highlights}
\end{table}

\begin{table*}[!h]
  \setlength{\tabcolsep}{3pt}         
  \renewcommand{\arraystretch}{0.8}   
  \small                              
  \caption{Experimental results of ablation study on the ACM dataset.}
  \vspace{-4mm}
  \label{tab: ablation study ACM}
  \begin{tabular}{@{}c|cccc@{}}
    \toprule
    Model & AUROC & AUPRC & F1-macro & Recall \\ \midrule
    \textbf{ChiGAD} & \textbf{0.9702} & \textbf{0.9518} & \textbf{0.9195} & \textbf{0.8947} \\ \midrule
    w/o Multi-Graph Chi-Square Filter & 0.9538 & 0.9236 & 0.8889 & 0.8499 \\
    w/o Interactive Meta-Graph Chi-Square Convolution & 0.8969 & 0.8410 & 0.7988 & 0.7343 \\
    w/o Contribution-Informed Cross-Entropy Loss & 0.9237 & 0.8449 & 0.8526 & 0.8070 \\
    Variant of ChiGAD (Beta filter) & 0.9411 & 0.9093 & 0.8735 & 0.8371 \\ \bottomrule
  \end{tabular}
\end{table*}

The statistics of datasets are summarized in Table~\ref{tab: datasets}. For heterogeneous datasets, the \#Features column in Table~\ref{tab: datasets} is an array listing the feature dimension for each node type; the anomaly ratio is defined as the ratio of abnormal nodes to target type nodes, rather than the ratio to all types of nodes.

\section{Implementation Setup}\label{}

For reproducibility, our code is available at \cite{Code}. Next, we explain how the data is split and how we set the hyper-parameters.

\subsection{Data Split}\label{app: data split}\label{}

The datasets are split with the open-sourced code published in GitHub repository\footnote{\texttt{https://github.com/squareRoot3/GADBench/tree/master}} of GADBench. Settings are detailed in Table \ref{tab: datasets}. 

\subsection{Hyper-Parameter Setting}\label{app: hyper-parameter setting}

Hyper-parameters for ChiGNN and ChiGAD experiments on homogeneous and heterogeneous graphs are summarized in 
Table  \ref{tab: hyper-param on homo G}
with performance evaluations in Tables \ref{tab: het - performance evaluation} and \ref{tab:homo-performance-evaluation-1}. The best model, selected via grid search based on validation set and F1-macro, is evaluated on the test set. Baselines use default hyper-parameters suggested by their authors. More generally, efficiency-oriented design is also being explored in other neural architectures, including phase mapping and sparsity-aware truncation for accelerated diffusion-model inference~\cite{zhao2026resilphase,zhao2026seeingendstepzero}.

\section{Proofs}

\subsection{Proof of Admissibility Condition} \label{sec: proof of admissibility condition}

Since \( f_i(w) \) is a real-valued function, we have \( |f_i(w)|^2 = f_i(w)^2 \). Thus, the integral becomes:
\begin{equation}\label{}
    \int_0^{\infty} \frac{f_i(w)^2}{w} \, dw. \nonumber
\end{equation}
Substituting the expression for \( f_i(w) \), we get:
\begin{align}\label{}
    \int_0^{\infty} \frac{1}{w} \left( \frac{1}{S_i} \frac{1}{2^{i} \Gamma(i)} \left(w (i + 1)\right)^{i - 1} e^{-\frac{w (i + 1)}{2}} \right)^2 \, dw. \nonumber
\end{align}
Simplifying the integrand:
\begin{align}\label{}
    \int_0^{\infty} \frac{1}{w} \left( \frac{1}{S_i} \frac{1}{2^{i} \Gamma(i)} \right)^2 \left(w (i + 1)\right)^{2(i - 1)} e^{-w (i + 1)} \, dw.\nonumber
\end{align}
Factoring out constants:
\begin{align}\label{}
    \left( \frac{1}{S_i} \frac{1}{2^{i} \Gamma(i)} \right)^2 (i + 1)^{2(i - 1)} \int_0^{\infty} w^{2(i - 1) - 1} e^{-w (i + 1)} \, dw.\nonumber
\end{align}
The integral is a Gamma function:
\begin{align} \label{}
    \int_0^{\infty} w^{2(i - 1) - 1} e^{-w (i + 1)} \, dw = \frac{\Gamma(2(i - 1))}{(i + 1)^{2(i - 1)}}.\nonumber
\end{align}
Substituting this back, we obtain:
\begin{align*} \label{}
    \left( \frac{1}{S_i} \frac{1}{2^{i} \Gamma(i)} \right)^2 (i + 1)^{2(i - 1)} \cdot \frac{\Gamma(2(i - 1))}{(i + 1)^{2(i - 1)}} = \left( \frac{1}{S_i} \frac{1}{2^{i} \Gamma(i)} \right)^2 \Gamma(2(i - 1)).
\end{align*}
Thus, we have
\begin{equation*}\label{}
    \int_0^{\infty} \frac{|f_i(w)|^2}{w} \, dw = \left( \frac{1}{S_i} \frac{1}{2^{i} \Gamma(i)} \right)^2 \Gamma(2(i - 1)) < \infty. 
\end{equation*}

So the Admissibility Condition is satisfied. {\hfill \qedsymbol}

\subsection{Empirical Proof of Chi-Square filters' Broad Coverage on Diverse Bands} 
\label{sec: broad coverage}

Empirical results in 
Table \ref{tab: Combined Expectations and Highlights}
show that the focus of $\{f_i\}$ broadly covers diverse bands over the full range of $[0,2]$.

\subsection{Proof of Theorem \ref{thm: 1}}\label{sec: proof for thm1}

We first prove the case of aligning 2 graph signals, and then naturally extend it to the case of $w$ signals. Supposing the input signals $\vect{x}_1,\vect{x}_2$ has been normalized, i.e. ${\vect{x}_{1}}^T\vect{x}_1={\vect{x}_{2}}^{\top}\vect{x}_2=1$, the aligned feature $\vect{x}_a$ is computed by the form of weighted sum as:
\begin{equation*}\label{}
    \vect{x}_a = w_1\vect{x}_1+w_2\vect{x}_2.
\end{equation*}
Let the high-frequency area of $\vect{x}_1,\vect{x}_2,\vect{x}_a$ be simply $S_1,S_2$ and $S_a$. We have
\begin{align}\label{eq: S_a}
    S_{a} &= \frac{\vect{x}_a^{\top}\vect{L}\vect{x}_a}{\vect{x}_a^{T}\vect{x}_a}
    = \frac{w_{1}^{2}S_{1}+w_{2}^2S_{2} +2w_{1}w_{2} \vect{x}_{1}^{\top}\vect{L}\vect{x}_{2}}{w_{1}^{2}+w_{2}^2 +2w_{1}w_{2} \vect{x}_{1}^{\top}\vect{x}_{2}}. 
\end{align}

Following the existing studies \cite{BWGNN,GCN,MGNN}, the graph spectral decomposition of the Laplacian matrix $L$ can be noted as 
\begin{equation*}
    \vect{L} = \vect{U}\vect{\Lambda} \vect{U}^{\top}.
\end{equation*}
Since $\vect{L}$ is a semi-positive definite matrix, the diagonal elements of $\vect{\Lambda}$ are all non-negative, and thereby so is $\vect{\Lambda}^{\frac{1}{2}}$. Based on this finding, it can be deduced that
\begin{equation*}
     \vect{L}^{\frac{1}{2}} = \vect{U}\vect{\Lambda}^{\frac{1}{2}}\vect{U}^{\top}
\end{equation*}
is also semi-positive definite. According to the Cauchy-Schwartz Theorem, it can be deduced that:
\begin{align*} 
    (\vect{x}_{1}^{\top}\vect{L}\vect{x}_{2})^{2}&=[(\vect{L}^{\frac{1}{2}}\cdot \vect{x}_{1})^{\top}(\vect{L}^{\frac{1}{2}}\cdot \vect{x}_{2})]^{2}\\
    &\leq \sqrt{[(\vect{L}^{\frac{1}{2}}\cdot \vect{x}_{1})^{\top}(\vect{L}^{\frac{1}{2}}\cdot \vect{x}_{1})]^2}\cdot \sqrt{[(\vect{L}^{\frac{1}{2}}\cdot \vect{x}_{2})^{\top}(\vect{L}^{\frac{1}{2}}\cdot \vect{x}_{2})]}\\
    &= |\vect{x}_{1}^{\top}\vect{L}\vect{x}_{1}| \cdot |\vect{x}_{2}^{\top}\vect{L}\vect{x}_{2}| = S_{1}S_{2},
\end{align*}
and thereby 
\begin{equation}\label{eq: cauchy-s 1}
    \vect{x}_1^{\top} \vect{L} \vect{x}_2 \in [-\sqrt{S_1S_2}, \sqrt{S_1S_2}].
\end{equation}
Similarly, it is easy to see that:
\begin{equation}\label{eq: cauchy-s 2}
    (\vect{x}_1^{\top} \vect{x}_2)^2 \leq |\vect{x}_1^{\top} \vect{x}_1||\vect{x}_2^{\top} \vect{x}_2| = 1,
\end{equation}
and thereby
\begin{equation*}
    \vect{x}_1^{\top} \vect{x}_2 \in [-1,1].
\end{equation*}
Combining Equations \ref{eq: S_a}, \ref{eq: cauchy-s 1} and \ref{eq: cauchy-s 2}, we can derive a lower bound of $S_a$ in the following two cases:

\noindent{\bf Case 1:} $w_1\cdot w_2\geq 0$. In this case, it can be inferred that
\begin{align*} \label{}
    S_a \geq \mathcal{LB}_{c-1} &:= \frac{w_1^2S_1+w_2^2S_2-2w_1w_2\sqrt{S_1S_2}}{w_{1}^{2}+w_{2}^2 +2w_{1}w_{2}}\\
    &= (\frac{w_1}{w_1+w_2}\sqrt{S_1}-\frac{w_2}{w_1+w_2}\sqrt{S_2})^2;
\end{align*}

\noindent{\bf Case 2:} $w_1\cdot w_2< 0$. In this case, it can be inferred that
\begin{align*} \label{}
    S_a \geq \mathcal{LB}_{c-2} &:= \frac{w_1^2S_1+w_2^2S_2+2w_1w_2\sqrt{S_1S_2}}{w_{1}^{2}+w_{2}^2 -2w_{1}w_{2}}\\
    &= (\frac{w_1}{w_1-w_2}\sqrt{S_1}+\frac{w_2}{w_1-w_2}\sqrt{S_2})^2.
\end{align*}

By symmetry, it can be obtained that $\mathcal{LB}_{c-1}=\mathcal{LB}_{c-2}$. Hence we only consider \textbf{Case 1}, where $w_1, w_2 \geq 0$. Let $t =\frac{w_1}{w_2}\geq 0$, then
\begin{align*}
    f(t) &:= \mathcal{LB}_{c-1} = \Big(\frac{t\sqrt{S_1}-\sqrt{S_2}}{t+1}\Big)^2,
\end{align*}
and 
\begin{align}\label{eq: f'}
    f^{'}(t) &= 2\Big(\frac{t\sqrt{S_1}-\sqrt{S_2}}{t+1}\Big)
    \frac{\sqrt{S_1}+\sqrt{S_2}}{(t+1)^2}.
\end{align}
Equation \ref{eq: f'} reveals that the lower bound $f(t)$ monotonically decreases on $[0,\sqrt{\frac{S_2}{S_1}}]$ where $f^{'}(t)\leq 0$ while monotonically increases on $(\sqrt{\frac{S_2}{S_1}}, +\infty)$ where $f^{'}(t) > 0$. With these findings, it is clear that
\begin{equation*}
    f(0) = S_2, ~\lim_{t\rightarrow +\infty}f(t)=S_1
\end{equation*}
serve as the potential upper bound of $f(t)$, and notably, they are both achievable or arbitrarily approximated by adjusting $t$, i.e. the weights $w_1, w_2$. Specifically, if $S_1\leq S_2$, the adjustment of  $w_2 \gg w_1 \geq 0$ deserves to be implemented, and vice versa ($w_1 \gg w_2 \geq 0$) if $S_1 > S_2$. To this end, the lower bound $f(t)$ of $S_a$ is sufficiently close to or even reaches an ideal upper bound $\max\{S_1,S_2\}$, i.e., 
\begin{equation*}
S_{high}(\vect{x}_a,\vect{L})=\max\big\{S_{high}(\vect{x}_i,\vect{L})\big\}_{i=1}^{2}    
\end{equation*}

Therefore, the proof for aligning two graph signals has been completed. Then, we extend this to $k\in\mathbb{Z}^+$ graph signals. Assuming the theorem holds 
with $k-1$ graph signals:
\begin{equation*}\label{}
S_{high}(\vect{x}_a,\vect{L})=\max\big\{S_{high}(\vect{x}_i,\vect{L})\big\}_{i=1}^{k-1}.    
\end{equation*}
Assume the following weighted sum $    \vect{x}_a = \sum_{i=1}^{k-1}w_i^{*}\vect{x}_i$

satisfies
\begin{equation*}\label{}
    S_{high}(\vect{x}_a,\vect{L}) = \max\big\{S_{high}(\vect{x}_i,\vect{L})\big\}_{i=1}^{k-1}.
\end{equation*}
Considering $\vect{x}_a$ as a new graph signal, the alignment of $\vect{x}_a$ and $\vect{x}_k$ degrades to the demonstrated case where there are only two graph signals. The aligned result of $\vect{x}_a, \vect{x}_k$, $\vect{x}_a^{'}$, can be inferred that:
\begin{align*} \label{}
    &S_{high}(\vect{x}_a^{'},\vect{L}) = \max\big\{S_{high}(\vect{x}_a,\vect{L}), S_{high}(\vect{x}_k,\vect{L})\big\}\\
    &= \max\big\{\max\big\{S_{high}(\vect{x}_i,\vect{L})\big\}_{i=1}^{k-1}, 
    S_{high}(\vect{x}_k,\vect{L})\big\}\\
    &= \max\big\{S_{high}(\vect{x}_i,\vect{L})\big\}_{i=1}^{k}.
\end{align*}
Since $\vect{x}_a^{'}$ acts as an linear alignment of all the $k$ graph signals $\vect{x}_1,~\vect{x}_2,\cdots,~\vect{x}_k$, Theorem \ref{thm: 1} is fully demonstrated. {\hfill \qedsymbol}

\section{Experimental Results of Ablation Study}\label{app: table of ablation study}
Ablation study for ChiGAD is shown in Table \ref{tab: ablation study ACM}. We can observe that ChiGAD outperforms its ablated variants on all metrics (AUROC, AUPRC, F1-macro, and Recall), indicating that each of its components contributes to improved anomaly detection. Notably, removing the Multi-Graph Chi-Square Filter or the Interactive Meta-Graph Convolution leads to the most substantial drop in performance, underscoring their critical role in capturing heterogeneous information and aligning feature dimensions across node types.

Excluding the Contribution-Informed Cross-Entropy Loss diminishes Recall, which shows that weighting anomalies based on their frequency contributions helps the model focus on challenging yet vital abnormal nodes. Furthermore, ChiGAD outperforms a Beta filter variant, suggesting that the Chi-Square filter’s unique properties are effective in identifying anomalies.

\end{document}